\documentclass[sigconf]{acmart}

\usepackage[utf8]{inputenc}
\usepackage[T1]{fontenc}

\usepackage{epstopdf}
\usepackage{caption}
\usepackage{subfigure}
\usepackage{algorithm}
\usepackage{algorithmic}
\usepackage{booktabs}
\usepackage{threeparttable}
\usepackage{multirow}
\usepackage{amssymb}
\usepackage{amsmath}
\usepackage{bm}
\usepackage{esvect}
\usepackage{array}
\usepackage{graphicx}
\AtBeginDocument{%
  \providecommand\BibTeX{{%
    \normalfont B\kern-0.5em{\scshape i\kern-0.25em b}\kern-0.8em\TeX}}}

\copyrightyear{2019}
\acmYear{2019}
\acmConference[MM '19]{Proceedings of the 27th ACM International Conference on Multimedia}{October 21--25, 2019}{Nice, France}
\acmBooktitle{Proceedings of the 27th ACM International Conference on Multimedia (MM '19), October 21--25, 2019, Nice, France}
\acmPrice{15.00}
\acmDOI{10.1145/3343031.3351042}
\acmISBN{978-1-4503-6889-6/19/10}

\begin{document}
\fancyhead{}
\title{SRINet: Learning Strictly Rotation-Invariant Representations for Point Cloud Classification and Segmentation}


\author{Xiao Sun}

\affiliation{%
  \institution{Institute of Computer Science and Technology, Peking University}
}
\email{xsun@pku.edu.cn}

\author{Zhouhui Lian}
\authornotemark[1]
\affiliation{%
  \institution{Institute of Computer Science and Technology, Peking University}
}
\email{lianzhouhui@pku.edu.cn}

\author{Jianguo Xiao}

\affiliation{%
  \institution{Institute of Computer Science and Technology, Peking University}
}
\email{xiaojianguo@pku.edu.cn}

%
%
%
%
%


\begin{abstract}
  Point cloud analysis has drawn broader attentions due to its increasing demands in various fields. Despite the impressive performance has been achieved on several databases, researchers neglect the fact that the orientation of those point cloud data is aligned. Varying the orientation of point cloud may lead to the degradation of performance, restricting the capacity of generalizing to real applications where the prior of orientation is often unknown. In this paper, we propose the point projection feature, which is invariant to the rotation of the input point cloud. A novel architecture is designed to mine features of different levels. We adopt a PointNet-based backbone to extract global feature for point cloud, and the graph aggregation operation to perceive local shape structure.  Besides, we introduce an efficient key point descriptor to assign each point with different response and help recognize the overall geometry. Mathematical analyses and experimental results demonstrate that the proposed method can extract strictly rotation-invariant representations for point cloud recognition and segmentation without data augmentation, and outperforms other state-of-the-art methods.
\end{abstract}

\ccsdesc[500]{Computing methodologies~Shape representations}
\ccsdesc[500]{Computing methodologies~Shape inference}
\ccsdesc[500]{Computing methodologies~Point-based models}

\keywords{point cloud, rotation invariance, 3D shape analysis}

\maketitle

\section{Introduction}
3D shape analysis attracts increasing attentions with the advancement of 3D sensors and computing resources. Extracting discriminative features from 3D models or scenes becomes demanding for its widespread applications, such as autonomous vehicles, robotics, and many other fields. As a fundamental format to represent 3D models, point cloud can be conveniently acquired by laser scanner, but is unsuitable to be fed into deep neural networks due to its uncertain point number and unordered permutation. Previous works usually convert a point cloud into voxels or a collection of views from multiple perspectives. Then, these regular data can be further processed by powerful convolution neural networks. Besides the time consumed in transformation, shape information is unavoidably missing since the newly-generated data can not cover all details of the original geometry.

\begin{figure*}[ht]
\centering
\includegraphics[height=2.5in]{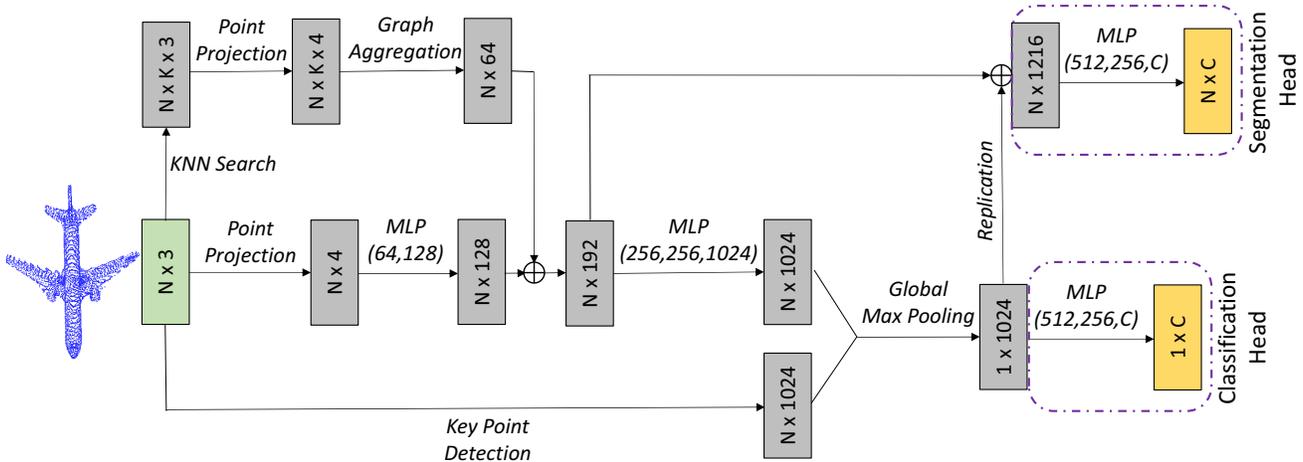}
\caption{The pipeline of SRINet. The proposed network begins with point projection operation to obtain rotation-invariant features. We use PointNet-based backbone to extract global information and apply graph aggregation to mine local shape feature. Key point response is explicitly defined according the normals of points in local regions to guide the network to better perceive the overall shape. Both classification and segmentation tasks share the same front head.}
\label{fig_pipeline}
\end{figure*}

\begin{figure}[ht]
\centering
\includegraphics[width=3.1in]{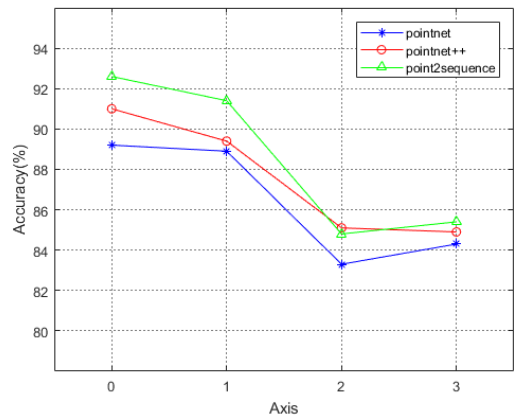}
\caption{Results when training with rotation augmentations.
Randomly rotating point cloud around different axes when training may lead to performance degradation in testing.}
\label{fig_2}
\end{figure}

Independent of networks that deal with regular data, point-based methods, like PointNet ~\cite{qi2017pointnet}, PointCNN ~\cite{li2018pointcnn} and PointSIFT ~\cite{jiang2018pointsift}, create a new paradigm to analyze unordered point cloud. Despite of the remarkable performance achieved, they limit themselves in processing the aligned 3D point clouds with canonical orientation. Even for the commonly used dataset, ModelNet ~\cite{wu20153d}, the orientations of the models are manually and roughly aligned, meaning that dealing with the problem of orientation bring in many inconveniences. Existing methods usually apply plenty of data augmentation to make the model robust to rotation. In our preliminary experiments, however, we find that rotating the point cloud along several axes may cause some degrees of degradation in performance (see Figure ~\ref{fig_2}). This demonstrates that the networks learn overmuch orientation information and obscure the intrinsic geometry characteristic.
PRIN ~\cite{you2018prin} translates the sparse points signal into voxel grids by Density-Aware Adaptive Sampling and employs Spherical Voxel Convolution to obtain approximate rotation-invariant features for every point. Such operations only improve the robustness to orientation, but can not achieve the goal of strict rotation invariance. PPF-FoldNet ~\cite{deng2018ppf} associates every point with a reference point and uses \textit{point pair feature} ~\cite{drost2010model} to substitute the original one. Despite of its strict rotation invariance, it is easy to construct different point pairs that are mapped into the same high-dimensional feature, which we believe impairs the capability of representations. In this paper, we propose the \textit{point projection feature}. Specifically, we select three main axes of the object and project every point to these axes to obtain a 3 dimensional feature, along with the norm of the vector starting from the central point to this point. The intuition is that the relative location relationship between points and the selected axes keeps fixed when rotating.
Thus, original 3D coordinates are mapped to the specially-designed 4D feature space, in which the representation of every point is rotation invariant to orientations. We use such mapping in two branches. In the main branch, we map the whole points to 4D space and feed the new representations to PointNet-based backbone to extract global features; in the side branch, we find $K$ nearest neighboring points for every point and apply graph aggregation operation to perceive local shape structure. Thus, the final encoded feature is independent of the orientation of input point cloud.
In addition, we hold the opinion that points in different position are of unequal importance for geometry perception. To be specific, corner points and edges are visually sensitive than those in flat regions. Automatically emphasizing such key points is essential for improving the quality of the obtained feature. Since there is no evidence for PointNet-like networks to automatically detect key points without extra supervision, we manually design a simple but effective key point detector to guide the neural network to focus on such key points.

Major contributions in this paper are threefold. First, we propose the point projection feature, a rotation-invariant representation that encodes the original coordinates of point cloud. Second, we design graph aggregation operation to mine local structure and explicitly introduce key point descriptor to emphasize the regions that are crucial for recognition. Third, extensive experiments demonstrate the superiority of our method in dealing with rotated point clouds.

\begin{figure}[h]
\centering
\includegraphics[width=3.1in]{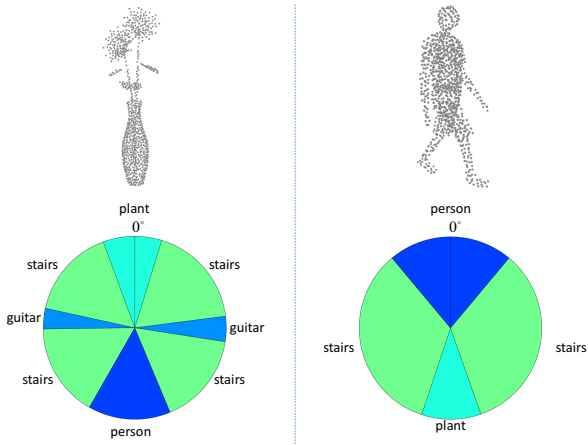}
\caption{Recognition results when rotating the upright objects by different angles.}
\label{fig_rot_360}
\end{figure}

\section{RELATED WORK }
\subsection{Deep learning on regular 3D data}
3D shape analysis relies on the quality of features extracted from 3D shapes. The appearance of large-scale 3D shape repositories and the development of hardware make it possible to leverage powerful deep networks to understand 3D data, and deep feature based methods outperform the traditional hand-crafted descriptors in most 3D vision tasks. Pioneer works ~\cite{maturana2015voxnet,wu20153d,qi2016volumetric} typically base on voxels for voxels are regularly arranged and suitable to feed into 3D convolution networks. However, 3D convolution occupies far more memories due to an extra dimension compared to 2D convolution, which limit the resolution of the voxels to be processed. In addition, 3D shape is perceived by its surface. Thus, operating on the elements inside the surface is a waste of computing resources. Another intuitive idea is to convert 3D shapes into a collection of views from multiple perspectives, then the proven techniques of 2D convolution can be adopted ~\cite{su2015multi,feng2018gvcnn}. View-based methods adopt view pooling to eliminate the order of views and become gradually robust to orientations as the number of views increases. Even so, they can only form a global descriptor, meaning that they can not carry out the delicate tasks like point labeling and matching.

\subsection{Deep learning on point cloud}
Different from other regular 3D representations, point cloud is difficult to mine local geometry and be processed by deep neural networks. Recently, researchers show increasing interest in point cloud processing ~\cite{hermosilla2018monte,yin2018p2p,roveri2018pointpronets}.
PointNet ~\cite{qi2017pointnet} is a pioneering work to study point cloud. The main idea of PointNet is to use point-wise convolution to map the original 3D coordinates to a high-dimension feature space, followed by a max pooling or average pooling operation to eliminate the effect of points permutation. Despite of its capacity to extract a global feature to represent point cloud, it neglects the exploring of forming local shape descriptors, making it hard to distinguish tiny diversity between contour-analogical 3D shapes. Several follow ups attempt to mine point cloud local structure in different ways. PointNet++ ~\cite{qi2017pointnet++} divides the whole point set into subsets and a simplified PointNet is applied repetitively for each of the subsets. These local features are grouped to make up a global representation. Due to the complicated process of division and grouping and repeated forward propagation, PointNet++ becomes time-consuming and sensitive to tuning, which often lead to worse results compared to PointNet in our preliminary experiments. PointCNN ~\cite{li2018pointcnn} proposes to find K nearest neighbor points for every point, from which a transformation matrix is learned to re-permutate these points, aiming to achieve permutation equivalence and perceive local regions. Since pre-multiplying original feature matrix by the learned transformation matrix is equal to swap or linearly recombine features in each row of original matrix, permutation equivalence can not be guaranteed. KCNet ~\cite{shen2018mining} proposes a shallow Kernel Correlation (KC) layer and K-NN Graph to incorporate local feature. We believe that deepening the KC layer may help mine more abstract and discriminative signals. All methods mentioned above need orientation-aligned point clouds as input, which limits their spread to practice when the prior of orientation is unknown. PRIN ~\cite{you2018prin} resorts to Spherical Voxel Convolution to extract features that are robust to orientations without data augmentation in the training process, but the performance degrades when testing with rotated data, indicating that it can not achieve strictly rotation-invariant representations.

\begin{figure}[t]
\centering
\subfigure[Three selected axes from point cloud.]{
  \includegraphics[height=1.5in]{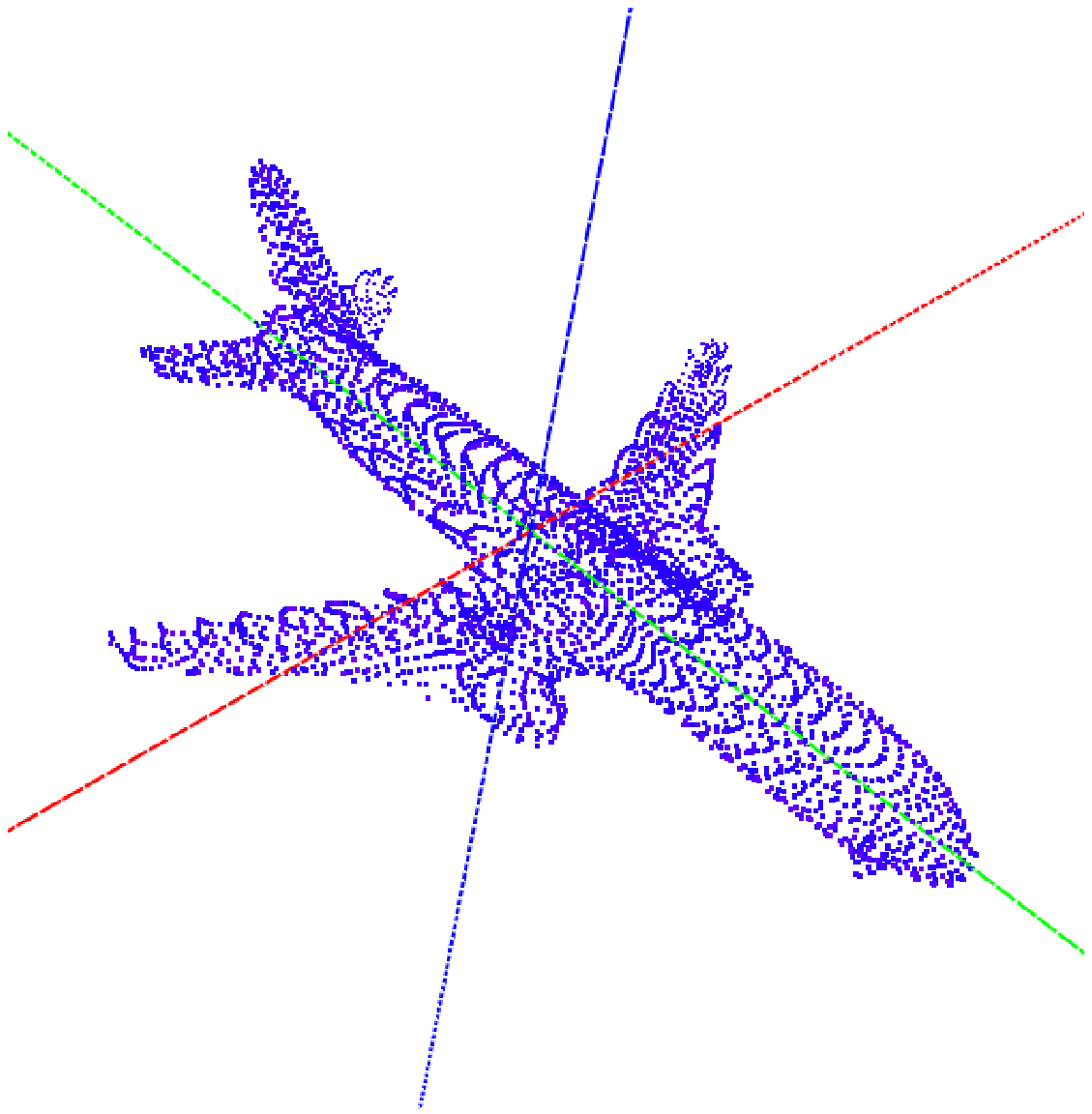}
  \label{fig_4_a}
  }
\centering
\hfill
\centering
\subfigure[Relation between original point and axes.]{
  \includegraphics[width=1.5in,height=1.5in]{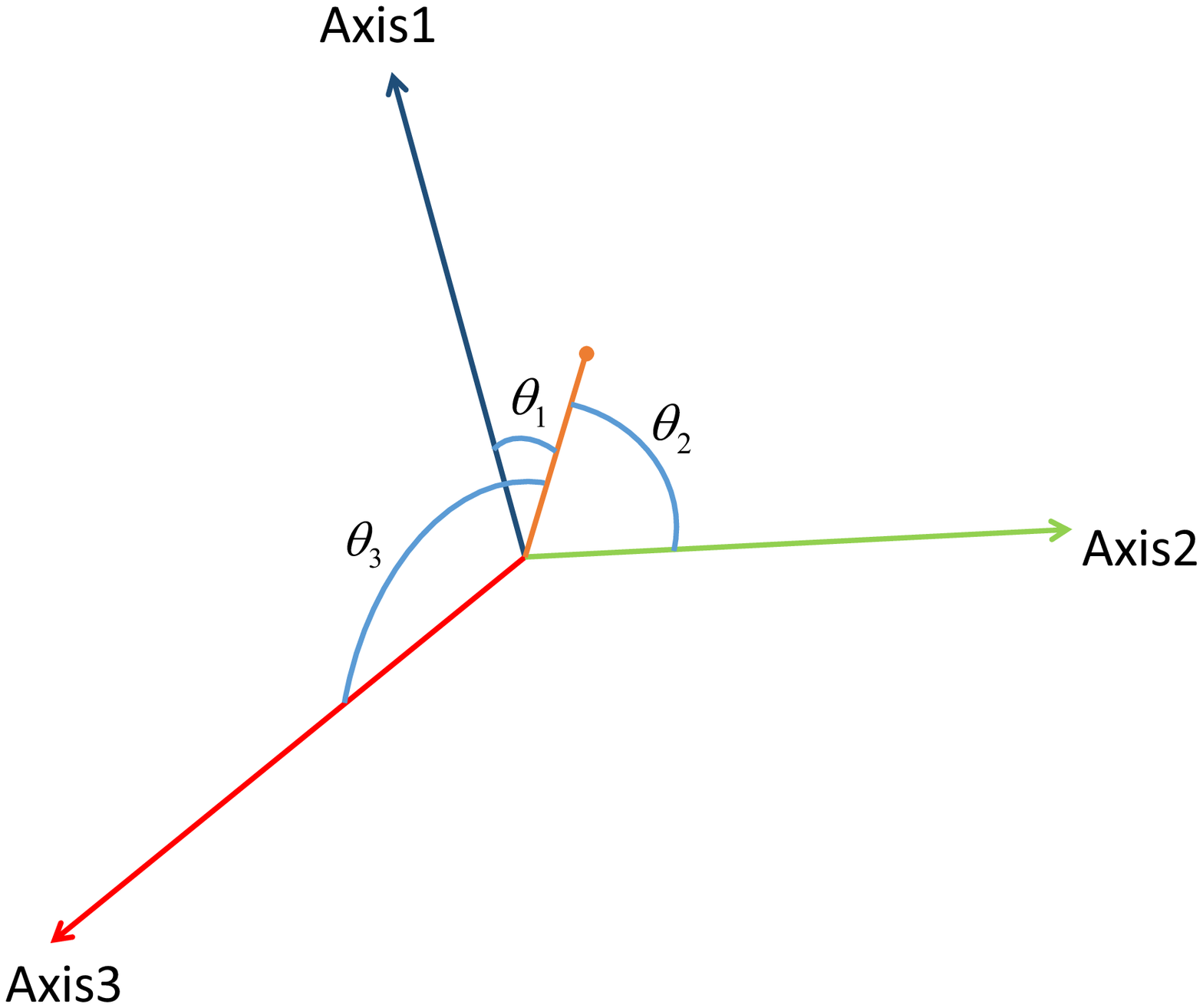}
  \label{fig_4_b}
  }
\caption{Illustration of point projection feature.}
\label{fig_ppf}
\end{figure}

\section{METHOD DESCRIPTION}
In this section, we introduce the proposed SRINet. The goal of our method is to obtain rotation-invariant representations for 3D point clouds, which can be used in the later applications ranging from classification to segmentation. We map the 3D coordinates into 4D point projection feature space and mine the features in both local and global receptive fields. We describe the details in the following subsections.

\subsection{Point Projection Feature}
Suppose the input point cloud is a set of points $X=\{x_i\in \mathbb{R}^3\}$ with random orientations, and we put the mass center at the origin.
The coordinates are also interpreted as the vectors starting from the origin. From the input vectors, we can freely choose three linearly independent axes (orthogonality is not necessary). Without loss of generality, we choose the vector with the maximum norm as axis 1 $(a1)$, the vector with the minimal norm as axis 2 $(a2)$, and the multiplication cross result of $a1$ and $a2$ as axis 3 $(a3)$. These three axes are scaled to unit norm. Clearly, no matter how the 3D object rotates, the relative location relationship between these axes and points keeps consistent. Then the original point cloud is encoded as a collection of point projection features
\begin{equation}
F_{\Omega} = \{f(a,x_1),f(a,x_2),...,f(a,x_N)\}\in \mathbb{R}^{N\times 4},
\end{equation}
where $a$ represents the three axes $(a_1,a_2,a_3)$ and $f$ denotes point projection mapping
\begin{equation}
f:(a,x_i)\to (\cos\langle a_1,x_i\rangle,\cos\langle a_2,x_i\rangle,\cos\langle a_3,x_i\rangle,|x_i|).
\end{equation}
We do not further calculate the angles between vectors because these patterns are difficult for neural networks to learn. Obviously, different points will not collide after mapping to 4 dimensional feature space.

\noindent\textbf{Proposition 1.} \textit{Point Projection Feature is invariant to the rotation of point cloud.}

\noindent\textbf{Proof.} Let us consider the point $x$ and three selected axes $a =(a_1,a_2,a_3)$, then part components of the 4D feature are remarked as
\begin{equation}
\langle x_n,a_1\rangle=f_1,\langle x_n,a_2\rangle=f_2,\langle x_n,a_3\rangle=f_3,
\end{equation}
where $x_n=x/|x|$. For simplicity, we denote $m_{ij}=\langle a_i,a_j\rangle$, and thus $m_{ij}=m_{ji}$. Then we can construct the matrix $M$ from the vector $C=(x_n,a)$:
\begin{equation}
\begin{bmatrix}x_n^T\\a_1^T\\a_2^T\\a_3^T\end{bmatrix}\begin{bmatrix}x_n&a_1&a_2&a_3\end{bmatrix}
=\begin{bmatrix}1&f_1&f_2&f_3\\f_1&1&m_{12}&m_{13}\\f_2&m_{21}&1&m_{23}\\f_3&m_{31}&m_{3}&1 \end{bmatrix}\triangleq M.
\end{equation}
Given matrix $M$, the vector $C$ can be obtained by applying singular value decomposition, $M=USV^T$, and $C=US^{1/2}V^{T}$. Note that, the axes are related to the values of $m_{ij}$ and the elements $m_{ij}$ keep fixed if the axes keep fixed. Rotating the original point cloud with orthogonal rotation matrix $R$ will not change the result matrix $M$: $(RC)^T(RC)=C^TR^TRC=C^TC=M$.

\noindent\textbf{Proposition 2.} \textit{Given the 4D point projection feature $f_i$ and 3 selected axes, the original point $x_i$ can be uniquely identified.}

\noindent\textbf{Proof.} By constructing the matrix $M$ and applying SVD, the coordinate of point $x_i$ can be easily calculated. Note that the particular solution $C^*$ is not up to a orthogonal matrix since the three axes are settled.



\begin{figure}[t]
\centering
\includegraphics[width=3in]{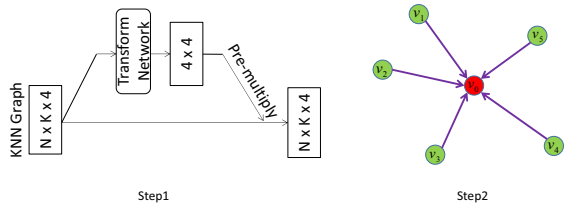}
\caption {Graph aggregation. A transform matrix is first learned and pre-multiply the feature matrix to aggregate features, then graph-based convolution and pooling are used for feature updates and fusing the local features into the central one.}
\label{fig_ga}
\end{figure}

\subsection{Local structure exploiting by graph aggregation}
In PointNet, transform matrix is learned from the whole points to integrate the features attached to every vertex, which is considered to impair the capability of perceiving local structure ~\cite{qi2017pointnet++,shen2018mining}.
Thus, it is necessary to extract features from local regions. Here we denote the local region as the K nearest neighboring points of a center point, and substract the coordinate of center point to neglect the relative position relationship to the whole point cloud. Intuitively, neighboring points construct local geometry structure together.
Mining local geometry features requires to exchange information between neighboring points and eliminate the effect of point permutation.
\begin{figure}[t]
\centering
\includegraphics[width=3.1in]{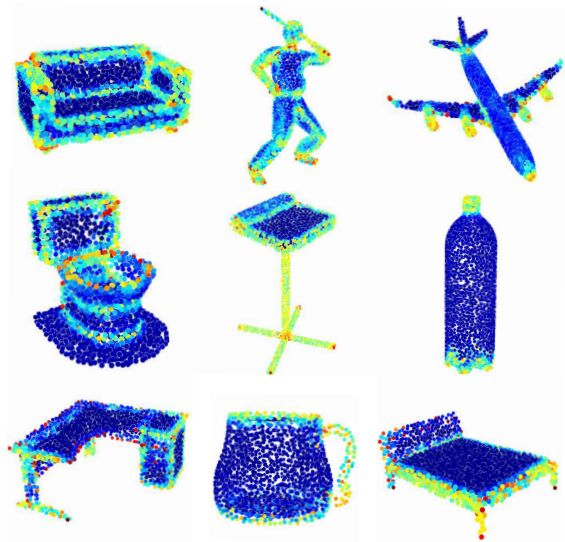}
\caption{Visualization of point clouds with the defined key point response values.}
\label{fig_kp}
\end{figure}
Only applying graph convolution among local points lacks interaction between features, and we conjecture that combining similar signatures may result in more salient ones. Inspired from PointNet, which learns a transform matrix and post-multiplies the feature matrix, we learn a similar transform matrix from local points and pre-multiply the feature matrix. Since we know, every row in feature matrix represents a feature vector attached to a point and pre-multiplying the feature matrix by transform matrix may linearly recombine the features. After that, graph convolution and pooling operations are used for feature update and fusion, which can be formulated as
\begin{equation}
f_i^{l+1}=P(F(f_{j\in N(i)}^{l})),
\end{equation}
where P denotes the pooling operation across neighboring points and F denotes graph-based convolution applied on each of the local points
\begin{equation}
h_i^{l+1}=w_{0}f_i^l+\sum_{j\in N_{center}/i}w_{1}f_j^l.
\end{equation}
Here, we update each of the local features using the points in the neighboring region around the center point, which is slightly different from the original definition of GCN ~\cite{kipf2016semi,bronstein2017geometric}.

\noindent Compared to EdgeConv operation proposed in ~\cite{wang2018dynamic}, which updates the local feature point by point, our newly generated feature of one point takes all points in local region into consideration. And we use max pooling to achieve permutation invariance and screen out the most salient signature among the local points. This aggregated signature reflects high-level abstract feature of local region
and can be concatenated with the global feature to form a complete point cloud representation.

\subsection{Key points detection}
In this context, key points denote the points lying on the edges or corners of the object. Exsiting works, such as ~\cite{liu2018point2sequence}, use attention module to highlight regions that are beneficial for recognition. In such data-driven methods, the degree of importance of each point is automatically learned without ground-truth key points for supervision, making it difficult to distinguish which point is truly important. Besides, it is hard to say the improvement of performance comes from the attention mechanism or from the increased number of parameters. We believe that more accurate information can be obtained by exploiting the intrinsic property of point cloud. The commonly used 3D corner detector, Harris 3D ~\cite{sipiran2011harris} achieves satisfying results, but is time-consuming and depends on parameter settings. It is universally acknowledged that the normals of points can reflect shape feature. Thus, we assign a response for every point by considering the changes of normals in its neighboring region
\begin{equation}
D_r=\sum_{i\in N(r)}\sin\langle n_i,n_r\rangle,
\end{equation}
where $n_i$ denotes the normal at point $x_i$.
Though simple, we find that it works well and the response of point cloud is visualized in Figure ~\ref{fig_kp}. High responses appear in the regions of edges, especially at the corners.
The calculated responses are integrated in the global representations of point cloud before global max pooling operation.

\subsection{Network Architecture}
The overall pipeline of the proposed method is demonstrated in Figure ~\ref{fig_pipeline}. The input point cloud is fed into two branches to extract both global and local features. Both of the two branches begin with point projection operation, mapping the 3D coordinates into 4 dimensional feature space. For the backbone, we use multilayer perceptron (MLP) to abstract pointwise feature. For the side branch, we leverage Graph Aggregation operation, which first learns a transform matrix from local points and pre-multiplies the feature matrix to recombine signatures, followed by graph convolutions and a max pooling layer to update features and form a local descriptor. The features from two branches are concatenated and then decorated with key point response values in two ways: pointwise multiplication or summation. We use global max pooling operation to eliminate the effect of point permutation and obtain a complete representation for point cloud. Classification and segmentation tasks share the the same representation of point cloud. In classification task, three extra fully-connected layers are used to serve as a classifier. In segmentation task, we replicate the representation and concatenate it with the features in previous layer, and then feed it to a three-layer MLP to produce scores for each point.

\section{EXPERIMENTS}
In this section, we validate the effectiveness of the proposed architecture in point cloud classification and part segmentation tasks, and conduct ablation study to evaluate the contribution of each components. SRINet is implemented with Tensorflow and runs on GTX1080Ti. We use Adam ~\cite{kingma2014adam} optimizer with initial learning rate 0.001 for training and decrease by 0.3 for 20 epochs in all experiments. For data augmentation, noise is added to perturb the object locations. We train the networks for 250 epochs to guarantee the convergence of model.

\subsection{Point Cloud Classification}
\textbf{Dataset.} We conduct classification experiments on ModelNet40 ~\cite{wu20153d}. The dataset consists of 12,311 CAD models from 40 categories, 9,843 of them are split for training and 2,468 for testing. Note that the orientation of these models are roughly aligned. We follow the same experimental settings as ~\cite{qi2017pointnet}. For each model, we uniformly sample 1024 points along with their normals as the network input.

Table \ref{table_1} compares the results of our method with several state-of-the-art works. NR/NR means not rotation of point clouds in both training and testing. NR/AR means training with no rotation augmentation and testing with arbitrary rotations. Our method gets the highest accuracy when testing with arbitrary rotations and outperforms other methods by a large margin. We also achieve comparable results compared to PointNet on non-rotation data. Besides, we achieve equal accuracy in rotation and non-rotation test settings, which means the obtained representation for point cloud is strictly rotation-invariant. PRIN degrades slightly when testing with rotations and shows strong robustness to rotations. Other works, however, fail to recognize the object with unseen orientations.

\begin{figure*}[ht]
\centering
\includegraphics[height=6in]{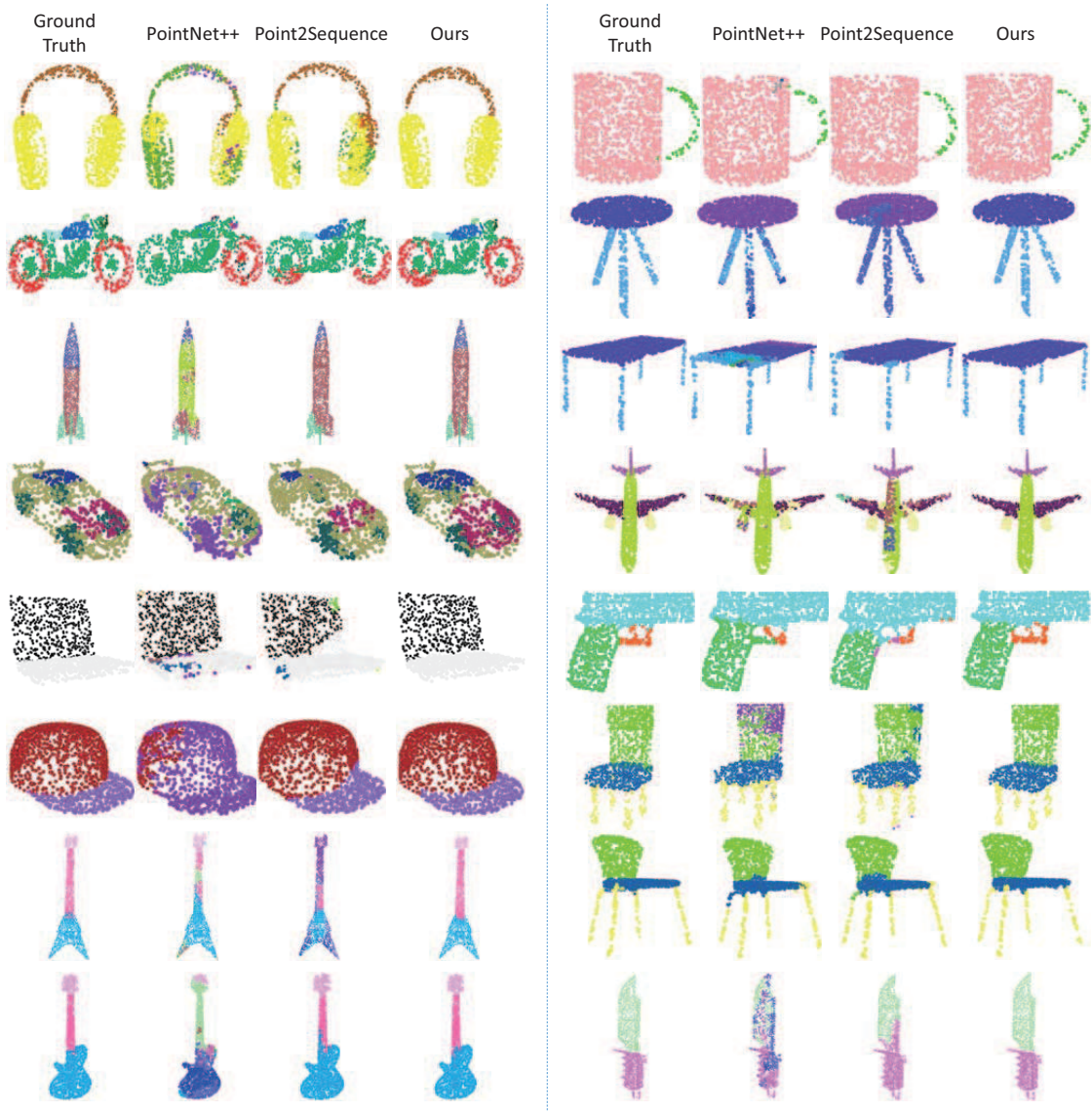}
\caption{Visualization of sampled point clouds after part segmentation.}
\label{fig_part_seg}
\end{figure*}

\begin{table}[t]
\newcommand{\tabincell}[2]{\begin{tabular}{@{}#1@{}}#2\end{tabular}}
\centering
\begin{tabular}{l|c|c} 
\hline
Method   &NR/NR   &NR/AR    \\
\hline
\multirow{1}*{\tabincell{c}{PointNet~\cite{qi2017pointnet}}}
    & 88.45         & 12.47     \\

\multirow{1}*{\tabincell{c}{PointNet++~\cite{qi2017pointnet++}}}
    & 89.42         & 21.35      \\

\multirow{1}*{\tabincell{c}{Point2Sequence~\cite{liu2018point2sequence}}}
    & \textbf{92.60}         & 10.53  \\

\multirow{1}*{\tabincell{c}{Kd-Network~\cite{klokov2017escape}}}
    & 86.20         & 8.49      \\

\multirow{1}*{\tabincell{c}{PRIN~\cite{you2018prin}}}
    & 80.13         & 69.85       \\
\hline
\multirow{1}*{\tabincell{c}{Ours}}
    & 87.01        & \textbf{87.01}       \\
\hline
  \end{tabular}
  \vspace{0.1cm}
  \caption{Comparison of different classification methods evaluated on ModelNet40 dataset.}
  \label{table_1}
\end{table}

\begin{table*}[t]
\newcommand{\tabincell}[2]{\begin{tabular}{@{}#1@{}}#2\end{tabular}}
\centering
\begin{tabular}{l|c|c|ccc|c} 
\hline
Method   &NR/NR   &NR/AR   &R$\times$10 &R$\times$20 &R$\times$30 &input size\\
\hline
\multirow{1}*{\tabincell{c}{PointNet~\cite{qi2017pointnet}}}
    & 93.42/83.43         & 45.66/28.26 & 61.02/41.59 &67.85/50.54 &74.91/58.66 & 2048$\times$3     \\

\multirow{1}*{\tabincell{c}{PointNet++~\cite{qi2017pointnet++}}}
    & 94.00/84.62         & 60.15/38.16 & 69.06/47.26 &70.01/49.26 &70.82/49.95 &1024$\times$3     \\

\multirow{1}*{\tabincell{c}{SyncSpecCNN~\cite{yi2017syncspeccnn}}}
    & 93.78/83.53         & 47.13/30.41 & 61.33/41.40 &68.10/50.76 &73.44/58.03 &2048$
    \times$33 \\

\multirow{1}*{\tabincell{c}{Kd-Network~\cite{klokov2017escape}}}
    & 90.33/82.36         & 40.66/24.76 & 59.11/38.70 &64.50/47.60 &69.33/51.06  &$2^{15}\times 3$    \\

\multirow{1}*{\tabincell{c}{PRIN~\cite{you2018prin}}}
    & 88.97/73.96         & 78.13/57.41 & 80.94/64.25 &83.83/67.68 &84.76/68.76   &2048$\times$ 3    \\
\hline
\multirow{1}*{\tabincell{c}{Ours}}
    &\multicolumn{5}{c|}{89.24/76.95} &2048$\times$ 3     \\
\hline
\end{tabular}

\vspace{0.1cm}
\caption{Quantitative segmentation results on ShapeNet part dataset.}
\label{table_part_seg}
\end{table*}
\subsection{Part Segmentation}
\textbf{Dataset.} We evaluate SRINet for part segmentation task on ShapeNet part dataset~\cite{yi2016scalable}. The dataset consists of 16,881 3D point cloud objects from 16 categories. The objects from various categories are segmented into 50 parts in total, and there are no overlap parts across different categories. For each object, a semantic label is assigned to every point. Each object contains no more than 5 parts. We use the processed dataset provided by ~\cite{qi2017pointnet} and randomly sample 2048 points with their normals from each object.

The rotation-invariant representations for point cloud can be also used for part segmentation task. We compare our work with PointNet~\cite{qi2017pointnet}, PointNet++~\cite{qi2017pointnet}, SyncSpecCNN~\cite{yi2017syncspeccnn}, Kd-Network~\cite{klokov2017escape} and PRIN~\cite{you2018prin}. We follow the same experimental settings as ~\cite{you2018prin} in evaluation, and three groups of settings are listed as follows:

1. Train and test with no rotations.

2. Train with no rotation augmentations and test with arbitrary rotations.

3. Train with 10/20/30 rotations for every model and test with arbitrary rotations.

The results are shown in Table ~\ref{table_part_seg}. State-of-the-art methods, such as PointNet, use orientation-aligned point clouds as their input and achieve good performance in the original task, but show great performance degradation when dealing with rotated point clouds. Training with increasing rotation augmentations may help improve the robustness to rotation, but the results are still worse than PRIN and ours. Besides, the improvement is obtained at a price of aggravating burden on computing resources. PRIN is not sensitive to rotations, however, it fails to achieve strictly rotation invariance. Our method neglects the effect of orientation and obtains best performance in segmenting rotated point cloud. The visualization of our segmentation results are demonstrated in Figure ~\ref{fig_part_seg}. We train these three models without rotation augmentation and rotate the input point clouds by a random angle when testing.

\begin{table}[ht]
\newcommand{\tabincell}[2]{\begin{tabular}{@{}#1@{}}#2\end{tabular}}
\centering
\begin{tabular}{cccc} 

\hline
Task   &Classification   &\multicolumn{2}{c}{Segmentation}    \\
Metric  &Acc(\%)  &Acc(\%)   &IoU(\%)   \\

\hline
Full & 87.01    & 89.24    & 76.95   \\
-GA   & 82.22    & 87.72    & 74.30   \\
-KPD   & 85.59    & 88.73    & 76.29   \\

\hline
  \end{tabular}
  \vspace{0.1cm}
  \caption{Quantitative results in ablation study.}
  \label{table_ablation_study}
\end{table}

\subsection{Ablation Study}
\noindent \textbf{Graph Aggregation}.
Graph aggregation operation is introduced in the side branch to exploit local geometry structure, and aggregate the features attached to the neighboring points around the center. We find it useful in point cloud recognition task, meaning that incorporating local structure helps perceive the global geometry. It also makes sense in segmentation task and improves the performance slightly. This is because precise segmentation requires global perception and local information only plays a secondary rule. The quantitative results for eliminating Graph Aggregation module are shown in Table ~\ref{table_ablation_study}.

\begin{table}[t]
\newcommand{\tabincell}[2]{\begin{tabular}{@{}#1@{}}#2\end{tabular}}
\centering
\begin{tabular}{cccc} 

\hline
\multirow{2}*{\tabincell{c}{Concatenation \\Methods}}  &Classification   &\multicolumn{2}{c}{Segmentation}    \\
  &Acc(\%)  &Acc(\%)   &IoU(\%)   \\

\hline
Multiplication & 85.26    & 88.81    & 75.52   \\
Summation   & 87.01    & 89.24    & 76.95   \\

\hline
  \end{tabular}
  \vspace{0.1cm}
  \caption{Results on different methods that incorporate key points response.}
  \label{table_kp_concat}
\end{table}

\noindent \textbf{Key Points Detection.} Intuitively, mining the skeleton and key points of an object is beneficial to recognize the whole shape. We directly define the key point response value instead of adopting learnable neural-network based attention mechanism. We combine the response values and global point cloud representations in two ways: multiplication and summation. As shown in Table ~\ref{table_kp_concat}, combination by summation is proved to be useful. However, combination by multiplication results in worse performance than that with no key point detection module.
And we remove the key points detection module to observe the effect to the whole model. The accuracy in classification experiment drops $1.42\%$ (from $87.01\%$ to $85.59\%$), and IoU value drops $0.66\%$ (from $76.95\%$ to $76.29\%$) without detecting key points. Though simple, this module brings in stable improvement for classification and segmentation tasks.

\subsection{The effect of parameters}
\noindent \textbf{Number of nearest neighboring points.}
We need to find K nearest neighboring points for each point in Graph Aggregation operation. From Table ~\ref{table_KNN}, we can see that the number of neighboring points is not crucial in classification, but greatly affects the segmentation task. As the number of neighboring points increases, the performance of segmentation keeps going up. We conjecture that segmentation relies on the receptive field of local region, and broader receptive field may lead to better perception of global shape.

\begin{table}[h]
\newcommand{\tabincell}[2]{\begin{tabular}{@{}#1@{}}#2\end{tabular}}
\centering
\scalebox{0.9}{
\begin{tabular}{ccccccc} 

\hline
\multicolumn{2}{l}{KNN point number}  & 16 & 25 & 36 & 49  & 64    \\
\hline
\multirow{1}*{\tabincell{c}{Classification}} & Acc(\%) & 86.85 &87.01 &86.93 & 86.89  &86.56\\
\hline
\multirow{2}*{\tabincell{c}{Segmentation}} & Acc(\%) & 88.19 &88.42 &88.89 & 89.00  &89.24\\
 & IoU(\%) & 74.96 &75.34 &76.19 & 76.56  &76.95 \\
\hline
  \end{tabular}}
  \vspace{0.1cm}
  \caption{The effect of the number of neighboring points that are found for every point.}
  \label{table_KNN}
\end{table}

\noindent \textbf{Number of input points.}
We vary the number of sampled points in the input point cloud to see if the proposed model is robust to the resolution of point cloud. The number ranges from 256 to 2048, shown in Table ~\ref{table_point_num}. We obtain the best results for both two tasks when the number of points is set to 1024. There is a tiny swing when going left or right from 1024, which suggests that SRINet is capable of extract valid local information despite of  the different distributions of local regions, and sampling 1024 points from the original point cloud is an optimum choice to cover the whole object.

\begin{table}[h]
\newcommand{\tabincell}[2]{\begin{tabular}{@{}#1@{}}#2\end{tabular}}
\centering
\begin{tabular}{cccccc} 

\hline
\multicolumn{2}{l}{Point Number}  & 256 & 512 & 1024 & 2048    \\
\hline
\multirow{1}*{\tabincell{c}{Classification}} & Acc(\%) & 85.87 &86.32 &87.01 & 85.83\\
\hline
\multirow{2}*{\tabincell{c}{Segmentation}} & Acc(\%) & 88.71 &88.97 &89.28 & 89.24\\
 & IoU(\%) & 77.07 &77.24 &77.28 & 76.95 \\
\hline
  \end{tabular}
  \vspace{0.1cm}
  \caption{The effect of the number of sampled point.}
  \label{table_point_num}
\end{table}

\subsection{Comparing with Point Pair Feature}
There exist several works that adopt point pair feature to reformulate the coordinates of point cloud and achieve strictly rotation invariance ~\cite{deng2018ppf,birdal2015point,birdal2017cad}. Here, we compare it with the proposed point projection feature. In preliminary experiments, we find it difficult for neural networks to extract discriminative patterns from the original point pair features, which incorporate calculating the angle between two defined vectors. Thus, we step back and replace the angles with their cosine values that can be calculated by the inner product of two normalized vectors. For fair comparison, we use PointNet architecture and conduct classification task on ModelNet40. The original 3D coordinates of point cloud are converted to 4D point pair features and point projection features respectively, and then fed to the network. The one with point projection features gets an accuracy of $82.2\%$, while the counterpart only gets $68.9\%$. This implies that the proposed point projection feature preserves more relative location information between points compared to point pair feature, and shows great superiority in terms of point cloud recognition.

\section{CONCLUSION}
In this paper, we proposed SRINet to extract the strictly rotation-invariant representation of point cloud. Point projection feature was introduced to reformulate the original 3D coordinates. We used graph aggregation to mine local structure and key point detection to guide the network to perceive the 3D shape. Experiments on classification and part segmentation tasks showed that our method outperforms other methods in dealing with rotated point clouds. In the future work, the choice of more stable axes needs to be further exploited to reduce the loss when converting the 3D coordinates to point projection features. Besides, how to better understand the point projection feature and generalize it to more applications is also an interesting work worth to be done in the future.

\section{Acknowledgments}
This work was supported by National Key Research and Development Program of China (2017YFB1002601), National Natural Science Foundation of China (Grant No.: 61672043 and 61672056) and Key Laboratory of Science, Technology and Standard in Press Industry (Key Laboratory of Intelligent Press Media Technology).

\bibliographystyle{ACM-Reference-Format}
\bibliography{sample-sigconf}

\end{document}